\documentclass[11pt,a4paper]{article}
\usepackage[hyperref]{acl2017}
\usepackage{times}
\usepackage{amsmath}
\usepackage{latexsym}
\usepackage{url}
\usepackage{graphicx}
\usepackage{subcaption}
\usepackage{multirow}
\usepackage{multicol}
\usepackage{booktabs}

\aclfinalcopy
\def\aclpaperid{19}

\newif\ifsubmit
\submittrue
\ifsubmit
\newcommand{\stnote}[1]{}
\newcommand{\ngnote}[1]{}
\newcommand{\lswnote}[1]{}
\newcommand{\sknote}[1]{}
\newcommand{\danote}[1]{}
\newcommand{\ecwnote}[1]{}

\else
\newcommand{\stnote}[1]{\textcolor{blue}{\textbf{ST: #1}}}
\newcommand{\ngnote}[1]{\textcolor{green}{\textbf{NG: #1}}}
\newcommand{\lswnote}[1]{\textcolor{violet}{\textbf{LSW: #1}}}
\newcommand{\sknote}[1]{\textcolor{red}{\textbf{SK: #1}}}
\newcommand{\danote}[1]{\textcolor{olive}{\textbf{DA: #1}}}
\newcommand{\ecwnote}[1]{\textcolor{purple}{\textbf{ECW: #1}}}

\fi

\title{A Tale of Two DRAGGNs: \\ A Hybrid Approach for Interpreting \\ Action-Oriented and Goal-Oriented Instructions}

\author{Siddharth Karamcheti, Edward C. Williams, Dilip Arumugam, \\
{\bf Mina Rhee}, {\bf Nakul Gopalan}, {\bf Lawson L.S. Wong}, {\bf Stefanie Tellex} \\
Department of Computer Science, Brown University, Providence, RI 02912 \\
\{{\tt siddharth\_karamcheti@}, {\tt edward\_c\_williams@}, {\tt dilip\_arumugam@}, \\ 
{\tt mina\_rhee@}, {\tt ngopalan@cs.}, {\tt lsw@}, {\tt stefie10@cs.}\}{\tt brown.edu}}

\begin{document}
\maketitle

\begin{abstract}

Robots operating alongside humans in diverse, stochastic environments must be able to accurately interpret natural language commands. These instructions often fall into one of two categories: those that specify a goal condition or target state, and those that specify explicit actions, or how to perform a given task.
Recent approaches have used reward functions as a semantic representation of goal-based commands, which allows for the use of a state-of-the-art planner to find a policy for the given task. However, these reward functions cannot be directly used to represent action-oriented commands. 
We introduce a new hybrid approach, the Deep Recurrent Action-Goal Grounding Network (DRAGGN), for task grounding and execution that handles natural language from either category as input, and generalizes to unseen environments.
Our robot-simulation results demonstrate that a system successfully interpreting both goal-oriented and action-oriented task specifications brings us closer to robust natural language understanding for human-robot interaction.
\end{abstract}

\section{Introduction}
\label{sec:intro}

\begin{figure}
\centering
	\includegraphics[width=.7\linewidth]{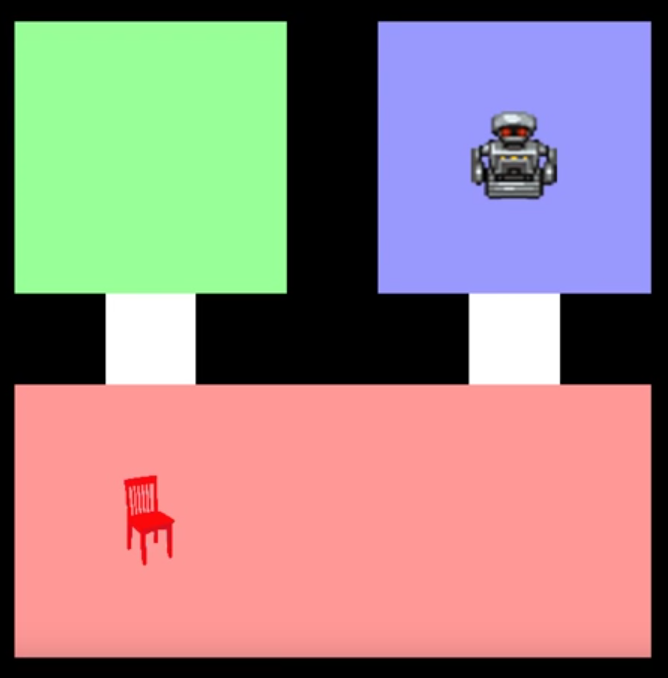}
	\caption{Sample configuration of the Cleanup World mobile-manipulator domain \citep{macglashan2015grounding}, used throughout this work. A possible goal-based instruction could be ``Take the chair to the green room,'' while a possible action-based instruction could be ``Go three steps south, then two steps west.''}
\label{fig:cleanup}
\end{figure}

 Natural language affords a convenient choice for delivering instructions to robots, as it offers flexibility, familiarity, and does not require users to have knowledge of low-level programming. In the context of grounding natural language instructions to tasks, human-robot instructions can be interpreted as either high-level goal specifications or low-level instructions for the robot to execute.

Goal-oriented commands define a particular target state specifying where a robot should end up, whereas action-oriented commands specify a particular sequence of actions to be executed.
For example, a human instructing a robot to ``go to the kitchen'' outlines a goal condition to check if the robot is in the kitchen. 
Alternatively, a human providing the command ``take three steps to the left'' defines a trajectory for the robot to execute.
We need to consider both forms of commands to understand the full space of natural language that humans may use to communicate their intent to robots. While humans also combine commands of both types into a single instruction, we make the simplifying assumption that a command belongs entirely to a single type and leave the task of handling mixtures and compositions to future work.

\begin{figure}
	\centering
   \includegraphics[width=\linewidth]{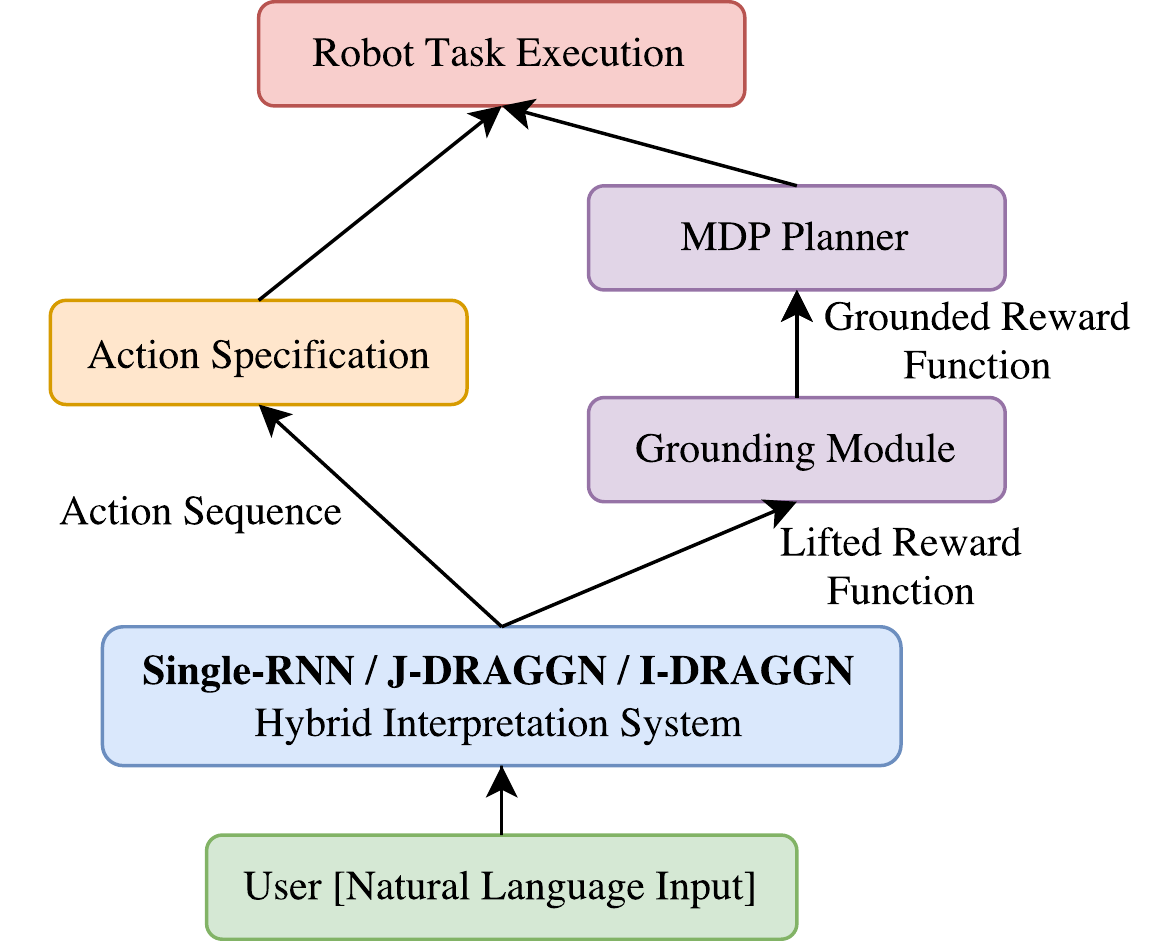}
    \caption{System for grounding both action-oriented (left branch) and goal-oriented (right branch) natural language instructions to executable robot tasks. Our main contribution is the hybrid interpretation system (blue box), for which we present two novel models based on the DRAGGN framework (J-DRAGGN and I-DRAGGN) in Section~\ref{sec:approach}.} 
    \label{fig:Pipeline}
\end{figure}
Existing approaches can be broadly divided into one of two regimes. Goal-based approaches like \citet{macglashan2015grounding} and \citet{Arumugam2017Accurate} leverage some intermediate task representation and then automatically find a low-level trajectory to achieve the goal using a planner. Other approaches, in the action-oriented regime, directly infer action sequences \citep{tellex11,matuszek2013learning,artzi13,AndreasK15} from the syntactic or semantic parse structure of natural language. However, these approaches can be computationally intractable for large state-action spaces or use ad-hoc methods to execute high-level language rather than relying on a planner. Furthermore, these methods are unable to adapt to dynamic changes in the environment; for example, consider an environment in which the wind, or some other force moves an object that a robot has been tasked with picking. Action sequence based approaches would fail to handle this without additional user input, while goal-based approaches would be able to re-plan on the fly, and complete the task.

To address the issue of dealing with both goal-oriented and action-oriented commands, we present a new language grounding framework that, given a natural language command, is capable of inferring the latent command type. Recent approaches leveraging deep neural networks have formulated the language grounding problem  as sequence-to-sequence learning or multi-label classification~\citep{MeiBW15,Arumugam2017Accurate}. Inspired by the recent success of neural networks to model programs that are highly compositional and sequential in nature, we present the Deep Recurrent Action/Goal Grounding Network (DRAGGN) framework, derived from the the Neural Programmer-Interpreter (NPI) of \citet{ReedF15}
and outlined in Section \ref{sec:draggn}. We introduce two instances of DRAGGN models, each with slightly different architectures. The first, the Joint-DRAGGN (J-DRAGGN) is defined in Section \ref{sec:cdraggn}, while the second, the Independent-DRAGGN (I-DRAGGN) is defined in Section \ref{sec:idraggn}.

\section{Related Work}
\label{sec:rw}

There has been a broad and diverse set of work examining how best to interpret and execute natural language instructions on a robot platform \citep{vogel10,tellex11,artzi13,howard2014natural,AndreasK15,HemachandraDHRS15,macglashan2015grounding,Paul2016abstract,MeiBW15,Arumugam2017Accurate}. \citet{vogel10} produce policies using language and expert trajectories based rewards, which allow for planning within a stochastic environment along with re-planning in case of failure.
\citep{tellex11} instead grounds language to trajectories satisfying the language specification. \cite{howard2014natural} chose to ground language to constraints given to an external planner, which is a much smaller space to perform inference over than trajectories. 
\citet{macglashan2015grounding} formulate language grounding as a machine translation problem, treating propositional logic functions as both a machine language and reward function. 
Reward functions or cost functions can allow richer descriptions of trajectories than plain constraints, as they can describe preferential paths.
Additionally, \citet{Arumugam2017Accurate} simplify the problem from one of machine translation to multi-class classification, learning a deep neural network to map arbitrary natural language instructions to the corresponding reward function.

Informing our distinction between action sequences and goal state representation is the division presented by \citet{dzifcak09}, who posited that natural language can be interpreted as \textit{both} a goal state specification and an action specification. Rather than producing both from each language command, our DRAGGN framework makes the simplifying assumption that only one representation captures the semantics of the language; additionally, our framework does not require a manually pre-specified grammar.

Recently, deep neural networks have found widespread success and application to a wide array of problems dealing with natural language \citep{Bengio2000ANP,Mikolov2010RecurrentNN,Mikolov2011ExtensionsOR,Cho2014LearningPR,Chung2014EmpiricalEO,Iyyer2015DeepUC}. Unsurprisingly, there have been some initial steps taken towards applying neural networks to language grounding problems.
\citet{MeiBW15} uses a recurrent neural network (RNN) with long short-term memory (LSTM) cells \citep{Hochreiter1997LongSM} to learn sequence-to-sequence mappings between natural language and robot actions. This model augments the standard sequence-to-sequence architecture by learning parameters that represent latent alignments between natural language tokens and robot actions. \citet{Arumugam2017Accurate} used an RNN-based model to produce grounded reward functions at multiple levels of an Abstract Markov Decision Process hierarchy \citep{Gopalan2016PlanningWA}, varying the abstraction level with the level of abstraction used in natural language. 

Our DRAGGN framework is closely related to the Neural Programmer-Interpreter (NPI) \citep{ReedF15}. The original NPI model is a controller trained via supervised learning to interpret and learn when to call specific programs/subprograms, which arguments to pass into the currently active program, and when to terminate execution of the current program. We draw a parallel between inferred NPI programs and our method of predicting either lifted reward functions or action trajectories.

\section{Problem Setting}
\label{sec:problem}

We consider the problem of mapping from natural language to robot actions within the context of Markov decision processes. A Markov decision process (MDP) is a five-tuple $\langle \mathcal{S}, \mathcal{A}, \mathcal{T}, \mathcal{R}, \gamma \rangle$ 
defining a state space $\mathcal{S}$, action space $\mathcal{A}$, state transition probabilities $\mathcal{T}$, reward function $\mathcal{R}$, and discount factor $\gamma$ \citep{Bellman1957,PutermanMarkovDP}. An MDP solver produces a policy that maps from states to actions in order to maximize the total expected discounted reward. 

While reward functions are flexible and expressive enough for a wide variety of task specifications, they are a brittle choice for specifying an exact sequence of actions, as enumerating every possible action sequence as a reward function (i.e. a specific reward function for the sequence Up 3, Down 2) can quickly become intractable. This paper introduces models that can produce desired behavior by inferring either reward functions or primitive actions. We assume that all available actions $\mathcal{A}$ and the full space of potential reward functions (\textit{i.e.}, the full space of possible tasks) are known \textit{a priori}. When a reward function is predicted by the model, an MDP planner is applied to derive the resultant policy (see system pipeline Figure~\ref{fig:Pipeline}).

We focus our evaluation of all models on the the Cleanup World mobile-manipulator domain \citep{macglashan2015grounding,Arumugam2017Accurate}. The Cleanup World domain consists of an agent in a $2$-D world with uniquely colored rooms and movable objects. A domain instance is shown in Figure~\ref{fig:cleanup}. The domain itself is implemented as an object-oriented Markov decision process (OO-MDP) where states are denoted entirely by collections of objects, with each object having its own identifier, type, and set of attributes \citep{Diuk2008AnOR}. Domain objects include rooms and interactable objects (\textit{e.g} a chair, basket, etc.) all of which have location and color attributes. Propositional logic functions can be used to identify relevant pieces of an OO-MDP state and their attributes; as in \citet{macglashan2015grounding} and \citet{Arumugam2017Accurate}, we treat these propositional functions as reward functions. In Figure~\ref{fig:cleanup}, the goal-oriented command ``take the chair to the green room'' may be represented with the reward function {\sf blockInRoom block0 room1}, where 
the {\sf blockInRoom} propositional function checks if the location attribute of {\sf block0} is contained in {\sf room1}.

\section{Approach}
\label{sec:approach}

We now outline the pipeline that converts natural language input to robot behavior. We begin by first defining the semantic task representation used by our grounding models that comes directly from the OO-MDP propositional functions of the domain. Next, we examine our novel DRAGGN framework for language grounding and, in particular, address the separate paths taken by action-oriented and goal-oriented commands through the system as seen in Figure \ref{fig:Pipeline}. Finally, we discuss two different implementations of the DRAGGN framework that make different assumptions about the relationship between tasks and constraints. Specifically, we introduce the Joint-DRAGGN (J-DRAGGN), that assumes a probabilistic dependence between tasks (i.e. {\sf goUp}) and the corresponding arguments (i.e. {\sf 5} steps) based on a natural language instruction, and the Independent-DRAGGN (I-DRAGGN) that treats tasks and arguments as independent given a natural language instruction.

\subsection{Semantic Representation}

\begin{table}[t]
	\centering
    \begin{small}
		\begin{tabular}{cc}
		\toprule
        Action-Oriented        &   Goal-Oriented   \\ \midrule
   {\sf goUp(numSteps)}   & {\sf agentInRoom(room)}  \\
   {\sf goDown(numSteps)} & {\sf blockInRoom(room)}  \\
   {\sf goLeft(numSteps)}  \\
   {\sf goRight(numSteps)} \\
        \bottomrule
  		\end{tabular}
    \end{small}
     \caption{Set of action-oriented and goal-oriented callable units that can be generated by our DRAGGN models in the Cleanup World domain.}

     \label{fig:functions}
\end{table}

In order to map arbitrary natural language instructions to either action trajectories or goal conditions, we require a compact but sufficiently expressive semantic representation for both. To this end, we define the \textit{callable unit}, which takes the form of a single-argument function. These functions are paired with \textit{binding arguments} whose possible values depend on the callable unit type.
As in \citet{macglashan2015grounding} and \citet{Arumugam2017Accurate}, our approach generates reward function templates, or \emph{lifted} reward functions, for goal-oriented tasks along with environment-specific constraints. Once these templates and constraints are resolved to get a grounded reward function, the associated goal-oriented tasks can be solved by an off-the-shelf planner thereby improving transfer and generalization capabilities.

Goal-oriented callable units (lifted reward functions) are paired with binding arguments that specify properties of environment entities that must be satisfied in order to achieve the goal. These binding arguments are later resolved by the Grounding Module (see Section \ref{sec:gm}) to produce grounded reward functions (OO-MDP propositional logic functions) that are handled by an MDP planner. 

Action-oriented callable units directly correspond to the primitive actions available to the robot and are paired with binding arguments defining the number of sequential executions of that action. The full set of callable units along with requisite binding arguments is shown in Table~\ref{fig:functions}.

\begin{figure*}[t]
  \centering
  \begin{subfigure}{.45\linewidth}
  	\centering
    \hspace*{-15pt}
    \includegraphics[height=.3\textheight]{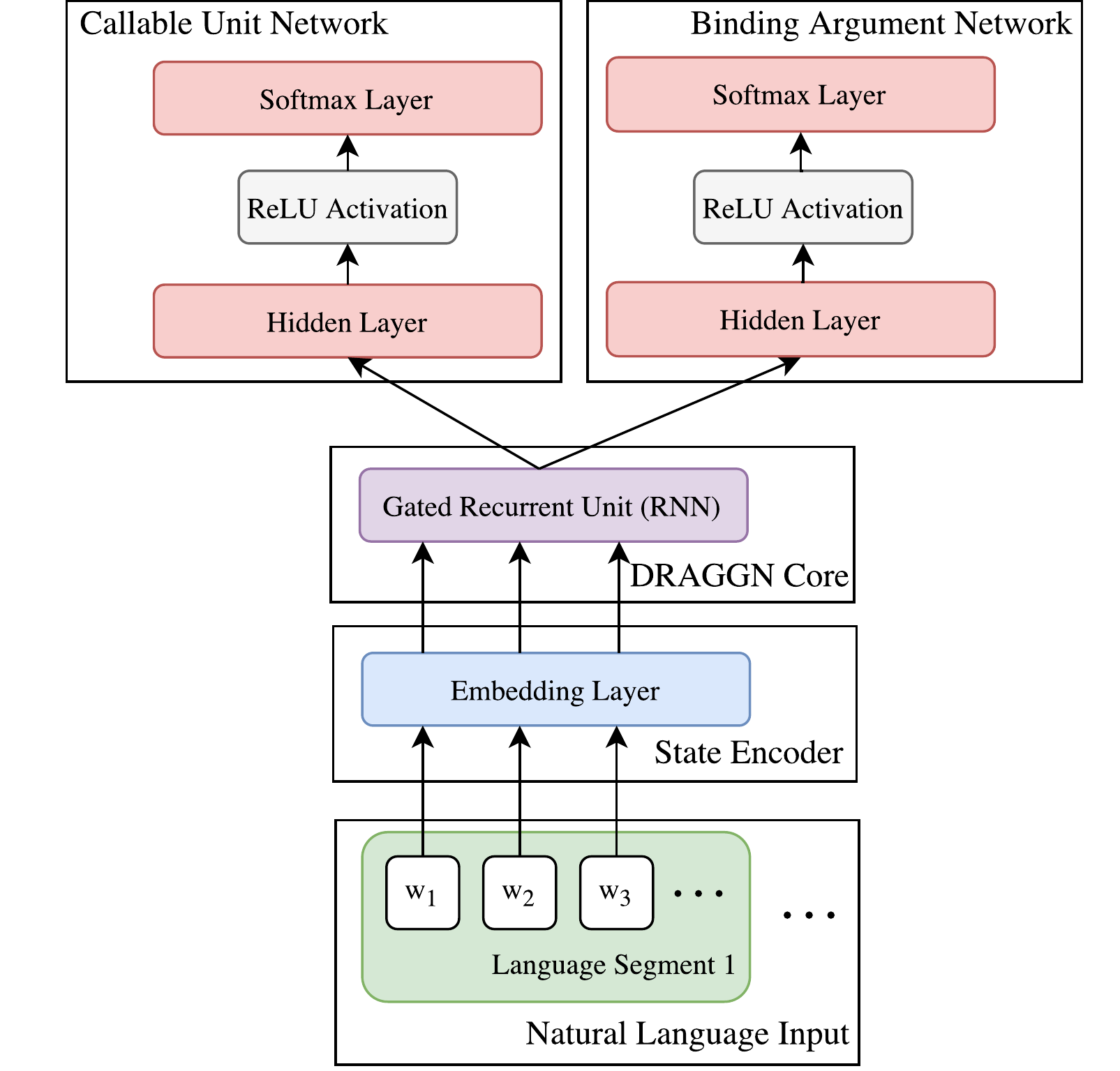}
 	\caption{Joint DRAGGN}
 	\label{fig:C-DRAGGN}
  \end{subfigure}
  \hfill
  \begin{subfigure}{.45\linewidth}
  	\centering
    \hspace*{-35pt}
    \includegraphics[height=.3\textheight]{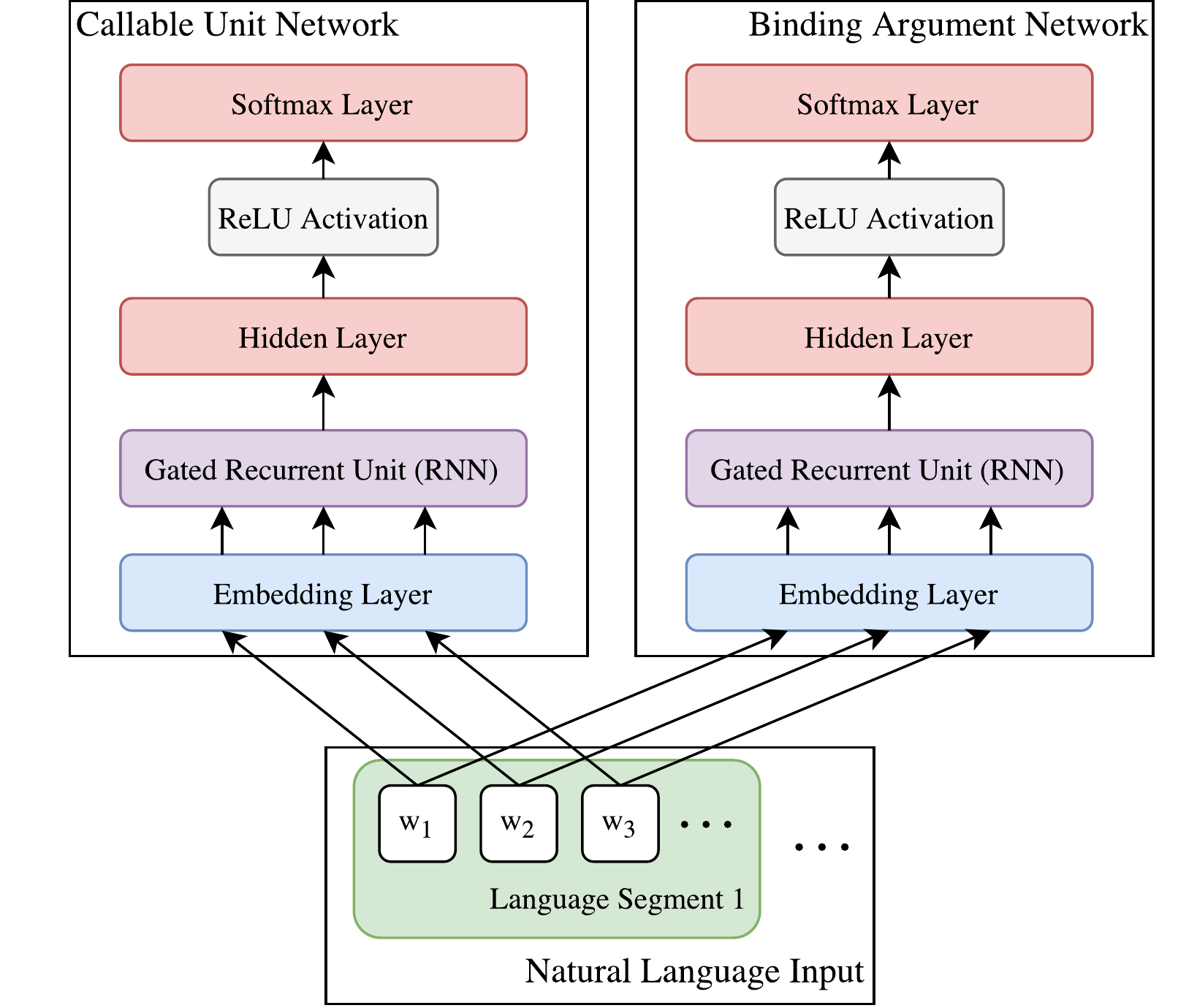}
 	\caption{Independent DRAGGN}
 	\label{fig:I-DRAGGN}
  \end{subfigure}
  \hfill
  \caption{Architecture diagrams for the two Deep Recurrent Action/Goal Grounding Network (DRAGGN) models, introduced in Sections~\ref{sec:cdraggn} and \ref{sec:idraggn}. Both architectures ground arbitrary natural language instructions to callable units (either actions or lifted reward functions), and binding arguments.  }
\end{figure*}

\subsection{Deep Recurrent Action/Goal Grounding Network (DRAGGN)}
\label{sec:draggn}

While the Single-RNN model of \citet{Arumugam2017Accurate} is effective, it cannot model the compositional argument structure of language. A unit-argument pair not observed at training time will not be predicted from input data, even if the constituent pieces were observed separately. Additionally, the Single-RNN model requires every possible unit-argument pair to be enumerated, to form the output space. As the environment grows to include more objects with richer attributes, this output space becomes intractable.

To resolve this, we introduce the Deep Recurrent Action/Goal Grounding Network (DRAGGN) framework. Unlike previous approaches, the DRAGGN framework maps natural language instructions to \emph{separate} distributions over callable units and (possibly multiple) binding constraints, generating either action sequences or goal conditions. By treating callable units and binding arguments as separate entities, we circumvent the combinatorial dependence on the size of the domain. 

This unit-argument separation is inspired by the Neural Programmer-Interpreter (NPI) of \citet{ReedF15}. The callable units output by DRAGGN are analogous to the subprograms output by NPI. Additionally, both NPI and DRAGGN allow for subprograms/callable units with an arbitrary number of arguments (by adding a corresponding number of Binding Argument Networks, as shown at the top right of Figure \ref{fig:C-DRAGGN}, each with its own output space).

We assume that each natural language instruction can be represented by a single unit-argument pair with only one argument. Consequently, in our experiments, we assume that sentences specifying sequences of commands have been segmented, and each segment is given to the model one at a time. The limitation to a single argument only arises because of the domain's simplicity; as mentioned above, it is straightforward to extend our models to handle extra arguments by adding extra Binding Argument Networks.

To formalize the DRAGGN objective, consider a natural language instruction $l$. Our goal is to find the callable unit $\hat{c}$ and binding arguments $\mathbf{\hat{a}}$ that maximize the following joint probability:
\begin{align}
	\label{eq:DRAGGN-Obj}
	\hat{c}, \hat{\mathbf{a}} &= \arg \max_{c, \mathbf{a}} \Pr(c, \mathbf{a} \mid l)
\end{align}

Depending on the assumptions made about the relationship between callable units $c$ and binding arguments $\mathbf{a}$, we can decompose the above objective in two ways: preserving the dependence between the two, and learning the relationship between the units and arguments jointly, and treating the two as independent. These two decompositions result in the Joint-DRAGGN and Independent-DRAGGN models respectively.

Given the training dataset of natural language and the space of unit-argument pairs, we train our DRAGGN models end-to-end by minimizing the sum of the cross-entropy losses between the predicted distributions and true labels for each separate distribution (\textit{i.e.} over callable units and binding arguments). At inference time, we first choose the callable unit with the highest probability given the natural language instruction. We then choose the binding argument(s) with highest probability from the set of valid arguments. The validity of a binding argument given a callable unit is given \textit{a priori}, by the specific environment, rather than being learned at training time. 

Our models were trained using Adam \citep{Kingma2014AdamAM}, for 125 epochs, with a batch size of 16, and a learning rate of 0.0001.

\subsection{Joint DRAGGN (J-DRAGGN)}
\label{sec:cdraggn}

The Joint DRAGGN (J-DRAGGN) models the joint probability in Equation~\ref{eq:DRAGGN-Obj}, coupled via the shared RNN state in the DRAGGN Core (as depicted in Figure \ref{fig:C-DRAGGN}), but selects the optimizer sequentially, as follows:
\begin{align}
	\hat{c}, \hat{\mathbf{a}} &= \arg \max_{c, \mathbf{a}} \Pr(c, \mathbf{a} \mid l) \\
    &\approx \arg \max_{\mathbf{a}} \, \left[ \arg \max_{c} \Pr(c, \mathbf{a} \mid l) \right] \nonumber
\end{align}

We first encode the constituent words of our natural language segment into fixed-size embedding vectors. From there, the sequence of word embeddings is fed through an RNN denoted by the DRAGGN Core\footnote{We use the gated recurrent unit (GRU) as our RNN cell, because of its effectiveness in natural language processing tasks, such as machine translation \citep{Cho2014LearningPR}, while requiring fewer parameters than the LSTM cell \citep{Hochreiter1997LongSM}.}. After processing the entire segment, the current gated recurrent unit (GRU) hidden state is then treated as a representative vector for the entire natural language segment. This single hidden core vector is then passed to both the Callable Unit Network and the Binding Argument Network, allowing for both networks to be trained jointly, enforcing a dependence between the two.

The Callable Unit Network is a two-layer feed-forward network using rectified linear unit (ReLU) activation. It takes the DRAGGN Core output vector as input to produce a softmax probability distribution over all possible callable units. The Binding Argument Network is a separate network with an identical architecture and takes the same input, but instead produces a probability distribution over all possible binding arguments. The two models do not need to share the same architecture; for example, callable units with multiple arguments require multiple different argument networks, one for each possible binding constraint.

\subsection{Independent DRAGGN (I-DRAGGN)}
\label{sec:idraggn}

The Independent DRAGGN (I-DRAGGN), contrary to the Joint DRAGGN, decomposes the objective from Equation~\ref{eq:DRAGGN-Obj} by treating callable units and binding arguments as being independent, given the original natural language instruction. More precisely, the I-DRAGGN objective is:
\begin{align}
	\hat{c}, \hat{\mathbf{a}} &= \arg \max_{c, \mathbf{a}} \Pr(c \mid l) \, \Pr(\mathbf{a} \mid l)
\end{align}

The I-DRAGGN network architecture is shown in Figure~\ref{fig:I-DRAGGN}.
Beyond the difference in objective functions, there is another key difference between the I-DRAGGN and J-DRAGGN architectures. Rather than encoding the constituent words of the natural language instruction once, and feeding the resulting embeddings through a DRAGGN Core to generate a shared core vector, the I-DRAGGN model embeds and encodes the natural language instruction \emph{twice}, using two separate embedding matrices and GRUs, one each for the callable unit and binding argument. In this way, the I-DRAGGN model encapsulates two disjoint neural networks, each with their own individual parameter sets that are trained independently.
The latter half of each individual network (the Callable Unit Network and Binding Argument Network) remains the same as that of the J-DRAGGN.

\subsection{Grounding Module}
\label{sec:gm}

If a goal-oriented callable unit is returned (\textit{i.e.} a lifted reward function), we require an additional step of completing the reward function with environment-specific variables. As described in \citet{Arumugam2017Accurate}, we use a Grounding Module to perform this step. The Grounding Module maps the inferred callable unit and binding argument(s) to a final grounded reward function that can be passed to an MDP planner. In our implementation, the Grounding Module is a lookup table mapping specific binding arguments to room ID tokens. A more advanced implementation of the Grounding Module would be required in order to handle domains with non-unique binding arguments (\textit{e.g.} resolving between multiple objects with overlapping attributes).

\section{Experiments}
\label{sec:experiments}

We assess the effectiveness of both our J-DRAGGN and I-DRAGGN models via instruction grounding accuracy for robot navigation and mobile-manipulation tasks. As a baseline, we compare against the state-of-the-art Single-RNN model introduced by \citet{Arumugam2017Accurate}.

\subsection{Procedure}
\label{sec:procedure}

\begin{table}[t]
	\centering
    \begin{small}
		\begin{tabular}{lcc}
		\toprule
        Natural Language &   Callable Unit  & Argument   \\ \midrule
        Go to the red room. & {\sf agentInRoom} & {\sf roomIsRed} \\
        Put the block in & {\sf blockInRoom} & {\sf roomIsGreen} \\
        \quad the green room. & & \\
        Go up three spaces. & {\sf goUp} & {\sf 3} \\
        \bottomrule
  		\end{tabular}
    \end{small}
     \caption{Examples of natural language phrases and corresponding callable units and arguments.}
     \label{fig:examples}
\end{table}

\begin{table*}
	\centering
    \begin{tabular}{lcccc}
    \toprule
    & Action-Oriented & Goal-Oriented & Action-Oriented (Unseen) & Overall \\ \midrule
    Single-RNN  & $95.8 \pm 0.1\%$  & $\mathit{87.2 \pm 0.9\%}$  & $0.0 + 0\%$ & $80.0 \pm 0.2\%$ \\
    J-DRAGGN    & $96.6 \pm 0.2\%$  & $\mathit{87.9 \pm 1.9\%}$ & $20.2 \pm 20.4\%$ & $83.7 \pm 2.8\%$ \\ 
	I-DRAGGN    & $\mathbf{97.0 \pm 0.2\%}$ & $84.9 \pm 1.8\%$  & $ \mathbf{97.0 + 0.0\%}$ & $\mathbf{94.7 \pm 0.5\%}$ \\
\bottomrule
\end{tabular}
	\caption{Action-oriented and goal-oriented accuracy results (mean and standard deviation across 3 random initializations) on both the standard and unseen datasets. \textbf{Bold} indicates the singular model that performed the best on the given task, whereas \textit{italics} denotes the best models that were within the margin of error of each other for the given task. The overall column was computed by taking an average of individual task accuracies, weighted by the number of test examples per task.}
     \label{table:singleResults}
\end{table*}

To conduct our evaluation, we use the dataset of natural language commands for the single instance of Cleanup World domain seen in Figure \ref{fig:cleanup}, from \citet{Arumugam2017Accurate}. In the user study, Amazon Mechanical Turk users were presented with trajectory demonstrations of a robot completing various navigation and object manipulation tasks. Users were prompted to provide natural language commands that they believed would have generated the observed behavior. Since the original dataset was compiled for analyzing the hierarchical nature of language, we were easily able to filter the commands down to only those using high-level goal specifications and low-level trajectory specifications. This resulted in a dataset of $3734$ natural language commands total.

To produce a dataset of action-specifying callable units, experts annotated low-level trajectory specifications from the \citet{Arumugam2017Accurate} dataset. For example, the command ``Down three paces, then up two paces, finally left four paces'' was segmented into ``down three spaces,'' ``then up two paces,''  ``finally left four paces,'' and was given a corresponding execution trace of {\sf goDown 3}, {\sf goUp 2}, {\sf goLeft 4}. The existing set of grounded reward functions in the dataset were converted to callable units and binding arguments. Examples of both types of language are presented in Table~\ref{fig:examples} with their corresponding callable unit and binding arguments.

To fully show the capabilities of our model, we tested on two separate versions of the dataset. The first is the standard dataset, consisting of a 90-10 split of the collected action-oriented and goal-oriented commands
We also evaluated our models on an ``unseen'' dataset, which consists of a specific train-test split that evaluates how well models can predict previously unseen action sequence combinations. For example, in this dataset the training data might consist only of action sequences of the form {\sf goUp 3}, and {\sf goDown 4}, while the test data would only consist of the ``unseen'' action sequence {\sf goUp 4}. Note that in both datasets, we assume that the test environment is configured the same as the train environment.

\subsection{Results}

Language grounding accuracies for our two DRAGGN models, as well as the baseline Single-RNN, are presented in Table~\ref{table:singleResults}.
All three models received the same set of training data, consisting of $2660$ low-level action-oriented segments and $693$ high-level goal-based sentences. All together, there are $17$ unique combinations action-oriented callable units and respective binding arguments, and $6$ unique combinations of goal-oriented callable units and binding arguments present in the data.
Then, we evaluated all three models on the same set of held-out data, which consisted of $295$ low-level segments and $86$ high-level sentences.

In aggregate, the models that use callable units for both action- and goal-based language grounding demonstrate superior performance to the Single-RNN baseline, largely due to their ability to generalize, and output combinations unseen at train time. We break down the performance on each task in the following three sections. 

\subsection{Action Prediction}

We evaluate the performance of our models on low-level language that directly specifies an action trajectory.
An instruction is correctly grounded if the output trajectory specification corresponds to the ground-truth action sequence.
To ensure fairness, we augment the output space of Single-RNN to include all distinct action trajectories found in the training data (an additional 17 classes, as mentioned previously).

All models perform generally well on this task, with Single-RNN correctly identifying the correct action callable unit on $95.8\%$ of test samples, while both DRAGGN models slightly outperform with on $96.6\%$ and $97.0\%$ respectively. 

\subsection{Goal Prediction}
\label{sec:GoalPred}

In addition to the action-oriented results, we evaluate the ability for each model to ground goal-based commands.
An instruction is correctly grounded if the output of the grounding module corresponds to the ground-truth (grounded) reward function.

In our domain, all models predict the correct grounded reward function with an accuracy of $84.9\%$ or higher, with the Single-RNN and J-DRAGGN models being too close to call.

\subsection{Unseen Action Prediction}
\label{sec:UnseenActionPred}

The Single-RNN baseline model is completely unable to produce unit-argument pairs that were never seen during training, whereas both DRAGGN models demonstrate some capacity for generalization. The I-DRAGGN model in particular demonstrates a strong understanding of each token within the original natural language utterances which, in large part, comes from the separate embedding spaces maintained for callable units and binding constraints respectively.

\section{Discussion}
\label{sec:discussion}

Our experiments show that the DRAGGN models have a clear advantage over the existing state-of-the-art in grounding action-oriented language. Furthermore, due to the factored nature of the output, I-DRAGGN generalizes well to unseen combinations of callable units and binding arguments.

Nevertheless, I-DRAGGN did not perform as well as Single-RNN and J-DRAGGN on goal-oriented language. This is possibly due to the small number of goal types in the dataset and the strong overlap in goal-oriented language. Whereas the Single-RNN and J-DRAGGN architectures may experience some positive transfer of information (due to the shared parameters in each of the two models), the I-DRAGGN model does not because of its assumed independence between callable units and binding arguments. This ability to allow for positive information transfer suggests that J-DRAGGN would perform best in environments where there is a strong overlap in the instructional language, with a relatively smaller but complex set of possible action sequences and goal conditions.

On action-oriented language, J-DRAGGN has grounding accuracy of around $20.2\%$ while I-DRAGGN achieves a near-perfect $97.0\%$. Since J-DRAGGN only encodes the input language instruction once, the resulting vector representation is forced to characterize both callable unit and binding argument features. While this can result in positive information transfer and improve grounding accuracy in some cases (\textit{e.g.} goal-based language), this enforced correlation heavily biases the model towards predicting combinations it has seen before. By learning separate representations for callable units and binding arguments,  I-DRAGGN is able to generalize significantly better. This suggests that I-DRAGGN would perform best in situations where the instructional language consists of many disjoint words and phrases.

While our results demonstrate that the DRAGGN framework is effective, more experimentation is needed to fully explore the possibilities and weaknesses of such models. One of the shortcomings in the DRAGGN models is the need for segmented data. We found that all evaluated models were unable to handle long, compositional instructions, such as ``Go up three steps, then down two steps, then left five steps''. Handling conjunctions of low-level commands requires extending our model to learn how to perform segmentation, or producing sequences of callable units and arguments.

\section{Conclusion}
\label{sec:conclusion}

In this paper, we presented the Deep Recurrent Action/Goal Grounding Network (DRAGGN), a hybrid approach that grounds natural language commands to either action sequences or goal conditions, depending on the language. We presented two separate neural network architectures that can accomplish this task, both of which factor the output space according to the compositional structure of our semantic representation. 

We show that overall the DRAGGN models significantly outperform the existing state of the art. Most notably, we show that the DRAGGN models are capable of generalizing to action sequences unseen during training time.

Despite these successes, there are still open challenges with grounding language to novel, unseen environment configurations. Furthermore, we hope to extend our models to handle 
instructions that are a mixture of goal-oriented and action-oriented language, as well as to long, sequential commands.
An instruction such as ``go to the blue room, but avoid going through the red hallway'' does not map to either an action sequence or a traditional, Markovian reward function. We believe new tools and approaches will need to be developed to handle such instructions, in order to handle the diversity and complexity of human natural language.
 
\section{Acknowledgements}

This material is based upon work supported by the National Science Foundation under grant number IIS-1637614 and the National Aeronautics and Space Administration under grant number NNX16AR61G.

Lawson L.S. Wong was supported by a Croucher Foundation Fellowship.

\bibliographystyle{acl_natbib}
\bibliography{references}

\begin{thebibliography}{}
\expandafter\ifx\csname natexlab\endcsname\relax\def\natexlab#1{#1}\fi

\bibitem[{Andreas and Klein(2015)}]{AndreasK15}
Jacob Andreas and Dan Klein. 2015.
\newblock Alignment-based compositional semantics for instruction following.
\newblock In {\em Conference on Empirical Methods in Natural Language
  Processing\/}.

\bibitem[{Artzi and Zettlemoyer(2013)}]{artzi13}
Yoav Artzi and Luke Zettlemoyer. 2013.
\newblock Weakly supervized learning of semantic parsers for mapping
  instructions to actions.
\newblock In {\em Annual Meeting of the Association for Computational
  Linguistics\/}.

\bibitem[{Arumugam et~al.(2017)Arumugam, Karamcheti, Gopalan, Wong, and
  Tellex}]{Arumugam2017Accurate}
Dilip Arumugam, Siddharth Karamcheti, Nakul Gopalan, Lawson~L.S. Wong, and
  Stefanie Tellex. 2017.
\newblock Accurately and efficiently interpreting human-robot instructions of
  varying granularities.
\newblock {\em CoRR\/} abs/1704.06616.

\bibitem[{Bellman(1957)}]{Bellman1957}
R.~Bellman. 1957.
\newblock A {M}arkovian decision process.
\newblock {\em Indiana University Mathematics Journal\/} 6:679--684.

\bibitem[{Bengio et~al.(2000)Bengio, Ducharme, Vincent, and
  Janvin}]{Bengio2000ANP}
Yoshua Bengio, R{\'e}jean Ducharme, Pascal Vincent, and Christian Janvin. 2000.
\newblock A neural probabilistic language model.
\newblock {\em Journal of Machine Learning Research\/} 3:1137--1155.

\bibitem[{Cho et~al.(2014)Cho, van Merrienboer, \c{C}aglar G{\'u}l\c{c}ehre,
  Bahdanau, Bougares, Schwenk, and Bengio}]{Cho2014LearningPR}
Kyunghyun Cho, Bart van Merrienboer, \c{C}aglar G{\'u}l\c{c}ehre, Dzmitry
  Bahdanau, Fethi Bougares, Holger Schwenk, and Yoshua Bengio. 2014.
\newblock Learning phrase representations using rnn encoder-decoder for
  statistical machine translation.
\newblock In {\em Empirical Methods in Natural Language Processing\/}.

\bibitem[{Chung et~al.(2014)Chung, \c{C}aglar G{\'u}l\c{c}ehre, Cho, and
  Bengio}]{Chung2014EmpiricalEO}
Junyoung Chung, \c{C}aglar G{\'u}l\c{c}ehre, Kyunghyun Cho, and Yoshua Bengio.
  2014.
\newblock Empirical evaluation of gated recurrent neural networks on sequence
  modeling.
\newblock {\em CoRR\/} abs/1412.3555.

\bibitem[{Diuk et~al.(2008)Diuk, Cohen, and Littman}]{Diuk2008AnOR}
Carlos Diuk, Andre Cohen, and Michael~L. Littman. 2008.
\newblock An object-oriented representation for efficient reinforcement
  learning.
\newblock In {\em International Conference on Machine Learning\/}.

\bibitem[{Dzifcak et~al.(2009)Dzifcak, Scheutz, Baral, and
  Schermerhorn}]{dzifcak09}
Juraj Dzifcak, Matthias Scheutz, Chitta Baral, and Paul Schermerhorn. 2009.
\newblock What to do and how to do it: Translating natural language directives
  into temporal and dynamic logic representation for goal management and action
  execution.
\newblock In {\em {IEEE} International Conference on Robotics and
  Automation\/}.

\bibitem[{Gopalan et~al.(2017)Gopalan, desJardins, Littman, MacGlashan, Squire,
  Tellex, Winder, and Wong}]{Gopalan2016PlanningWA}
Nakul Gopalan, Marie desJardins, Michael~L. Littman, James MacGlashan, Shawn
  Squire, Stefanie Tellex, John Winder, and Lawson~L.S. Wong. 2017.
\newblock Planning with abstract {M}arkov decision processes.
\newblock In {\em International Conference on Automated Scheduling and
  Planning\/}.

\bibitem[{Hemachandra et~al.(2015)Hemachandra, Duvallet, Howard, Roy, Stentz,
  and Walter}]{HemachandraDHRS15}
Sachithra Hemachandra, Felix Duvallet, Thomas~M. Howard, Nicholas Roy, Anthony
  Stentz, and Matthew~R. Walter. 2015.
\newblock Learning models for following natural language directions in unknown
  environments.
\newblock In {\em {IEEE} International Conference on Robotics and
  Automation\/}.

\bibitem[{Hochreiter and Schmidhuber(1997)}]{Hochreiter1997LongSM}
Sepp Hochreiter and J\"{u}rgen Schmidhuber. 1997.
\newblock Long short-term memory.
\newblock {\em Neural Computation\/} 9:1735--1780.

\bibitem[{Howard et~al.(2014)Howard, Tellex, and Roy}]{howard2014natural}
Thomas~M. Howard, Stefanie Tellex, and Nicholas Roy. 2014.
\newblock A natural language planner interface for mobile manipulators.
\newblock In {\em IEEE International Conference on Robotics and Automation\/}.

\bibitem[{Iyyer et~al.(2015)Iyyer, Manjunatha, Boyd-Graber, and
  Daum{\'e}}]{Iyyer2015DeepUC}
Mohit Iyyer, Varun Manjunatha, Jordan~L. Boyd-Graber, and Hal Daum{\'e}. 2015.
\newblock Deep unordered composition rivals syntactic methods for text
  classification.
\newblock In {\em Conference of the Association for Computational
  Linguistics\/}.

\bibitem[{Kingma and Ba(2014)}]{Kingma2014AdamAM}
Diederik~P. Kingma and Jimmy Ba. 2014.
\newblock Adam: A method for stochastic optimization.
\newblock {\em CoRR\/} abs/1412.6980.

\bibitem[{MacGlashan et~al.(2015)MacGlashan, Babe{\c s}-Vroman, desJardins,
  Littman, Muresan, Squire, Tellex, Arumugam, and
  Yang}]{macglashan2015grounding}
James MacGlashan, Monica Babe{\c s}-Vroman, Marie desJardins, Michael~L.
  Littman, Smaranda Muresan, Shawn Squire, Stefanie Tellex, Dilip Arumugam, and
  Lei Yang. 2015.
\newblock Grounding english commands to reward functions.
\newblock In {\em Robotics: Science and Systems\/}.

\bibitem[{Matuszek et~al.(2012)Matuszek, Herbst, Zettlemoyer, and
  Fox}]{matuszek2013learning}
Cynthia Matuszek, Evan Herbst, Luke Zettlemoyer, and Dieter Fox. 2012.
\newblock Learning to parse natural language commands to a robot control
  system.
\newblock In {\em International Symposium on Experimental Robotics\/}.

\bibitem[{Mei et~al.(2016)Mei, Bansal, and Walter}]{MeiBW15}
Hongyuan Mei, Mohit Bansal, and Matthew~R. Walter. 2016.
\newblock Listen, attend, and walk: Neural mapping of navigational instructions
  to action sequences.
\newblock In {\em AAAI Conference on Artificial Intelligence\/}.

\bibitem[{Mikolov et~al.(2010)Mikolov, Karafi{\'a}t, Burget, Cernock{\'y}, and
  Khudanpur}]{Mikolov2010RecurrentNN}
Tomas Mikolov, Martin Karafi{\'a}t, Luk{\'a}s Burget, Jan Cernock{\'y}, and
  Sanjeev Khudanpur. 2010.
\newblock Recurrent neural network based language model.
\newblock In {\em Interspeech\/}.

\bibitem[{Mikolov et~al.(2011)Mikolov, Kombrink, Burget, Cernock{\'y}, and
  Khudanpur}]{Mikolov2011ExtensionsOR}
Tomas Mikolov, Stefan Kombrink, Luk{\'a}s Burget, Jan Cernock{\'y}, and Sanjeev
  Khudanpur. 2011.
\newblock Extensions of recurrent neural network language model.
\newblock In {\em IEEE International Conference on Acoustics, Speech, and
  Signal Processing\/}.

\bibitem[{Paul et~al.(2016)Paul, Arkin, Roy, and Howard}]{Paul2016abstract}
Rohan Paul, Jacob Arkin, Nicholas Roy, and Thomas~M. Howard. 2016.
\newblock Efficient grounding of abstract spatial concepts for natural language
  interaction with robot manipulators.
\newblock In {\em Robotics: Science and Systems\/}.

\bibitem[{Puterman(1994)}]{PutermanMarkovDP}
Martin~L. Puterman. 1994.
\newblock Markov decision processes: Discrete stochastic dynamic programming.

\bibitem[{Reed and de~Freitas(2016)}]{ReedF15}
Scott~E. Reed and Nando de~Freitas. 2016.
\newblock Neural programmer-interpreters.
\newblock In {\em International Conference on Learning Representations\/}.

\bibitem[{Tellex et~al.(2011)Tellex, Kollar, Dickerson, Walter, Banerjee,
  Teller, and Roy}]{tellex11}
Stefanie Tellex, Thomas Kollar, Steven Dickerson, Matthew~R. Walter,
  Ashis~Gopal Banerjee, Seth Teller, and Nicholas Roy. 2011.
\newblock Understanding natural language commands for robotic navigation and
  mobile manipulation.
\newblock In {\em AAAI Conference on Artificial Intelligence\/}.

\bibitem[{Vogel and Jurafsky(2010)}]{vogel10}
Adam Vogel and Dan Jurafsky. 2010.
\newblock Learning to follow navigational directions.
\newblock In {\em Annual Meeting of the Association for Computational
  Linguistics\/}.

\end{thebibliography}

\end{document}